\documentclass[11pt,a4paper]{article}
\usepackage[hyperref]{emnlp-ijcnlp-2019}
\usepackage{times}
\usepackage{latexsym}
\usepackage{amsfonts}
\usepackage{latexsym}
\usepackage{bookmark}
\usepackage{microtype}
\usepackage{subfigure}
\usepackage{bbm}
\usepackage{amsmath}
\usepackage{graphicx}
\usepackage{url}
\usepackage{amssymb}
\usepackage{enumitem}
\usepackage{amsbsy}
\usepackage{booktabs}

\aclfinalcopy 

\title{Are You for Real? Detecting Identity Fraud via Dialogue Interactions}

\author{Weikang Wang$^{1,2}$, 
	Jiajun Zhang$^{1,2}$, 
	Qian Li$^3$, 
	Chengqing Zong$^{1,2,4}$ \and 
	Zhifei Li$^3$ \\
	$^1$ National Laboratory of Pattern Recognition, Institute of Automation, CAS, Beijing, China \\ 
	$^2$ University of Chinese Academy of Sciences, Beijing, China \\
	$^3$ Mobvoi, Beijing, China \\
	$^4$ CAS Center for Excellence in Brain Science and Intelligence Technology, Beijing, China  \\
	{\tt \{weikang.wang, jjzhang, cqzong\}@nlpr.ia.ac.cn} \\ 
	{\tt \{qli, zfli\}@mobvoi.com}}

\date{}

\begin{document}
\maketitle

\begin{abstract}
	Identity fraud detection is of great importance in many real-world scenarios such as the financial industry. However, few studies addressed this problem before. In this paper, we focus on identity fraud detection in loan applications and propose to solve this problem with a novel interactive dialogue system which consists of two modules. One is the knowledge graph (KG) constructor organizing the personal information for each loan applicant. The other is structured dialogue management that can dynamically generate a series of questions based on the personal KG to ask the applicants and determine their identity states. We also present a heuristic user simulator based on problem analysis to evaluate our method. Experiments have shown that the trainable dialogue system can effectively detect fraudsters, and achieve higher recognition accuracy compared with rule-based systems. Furthermore, our learned dialogue strategies are interpretable and flexible, which can help promote real-world applications.\footnote{\url{https://github.com/Leechikara/Dialogue-Based-Anti-Fraud}}
\end{abstract}

\section{Introduction}
Identity fraud is one person using another person's personal information or combining a few pieces of real data with bogus information to deceive a third person. Nowadays, identity fraud is becoming an increasingly prevalent issue and has left many financial firms nursing huge losses. Besides, for persons whose identities have been stolen, they may receive unexpected bills and their credit will also be affected. Although identity fraud is a very serious problem in modern society, there are no effective fraud detection methods at present and little attention has been paid to this problem.

\begin{figure}[!t]
	\centering
	\setlength{\abovecaptionskip}{0.2cm}
	\includegraphics[width=0.48\textwidth]{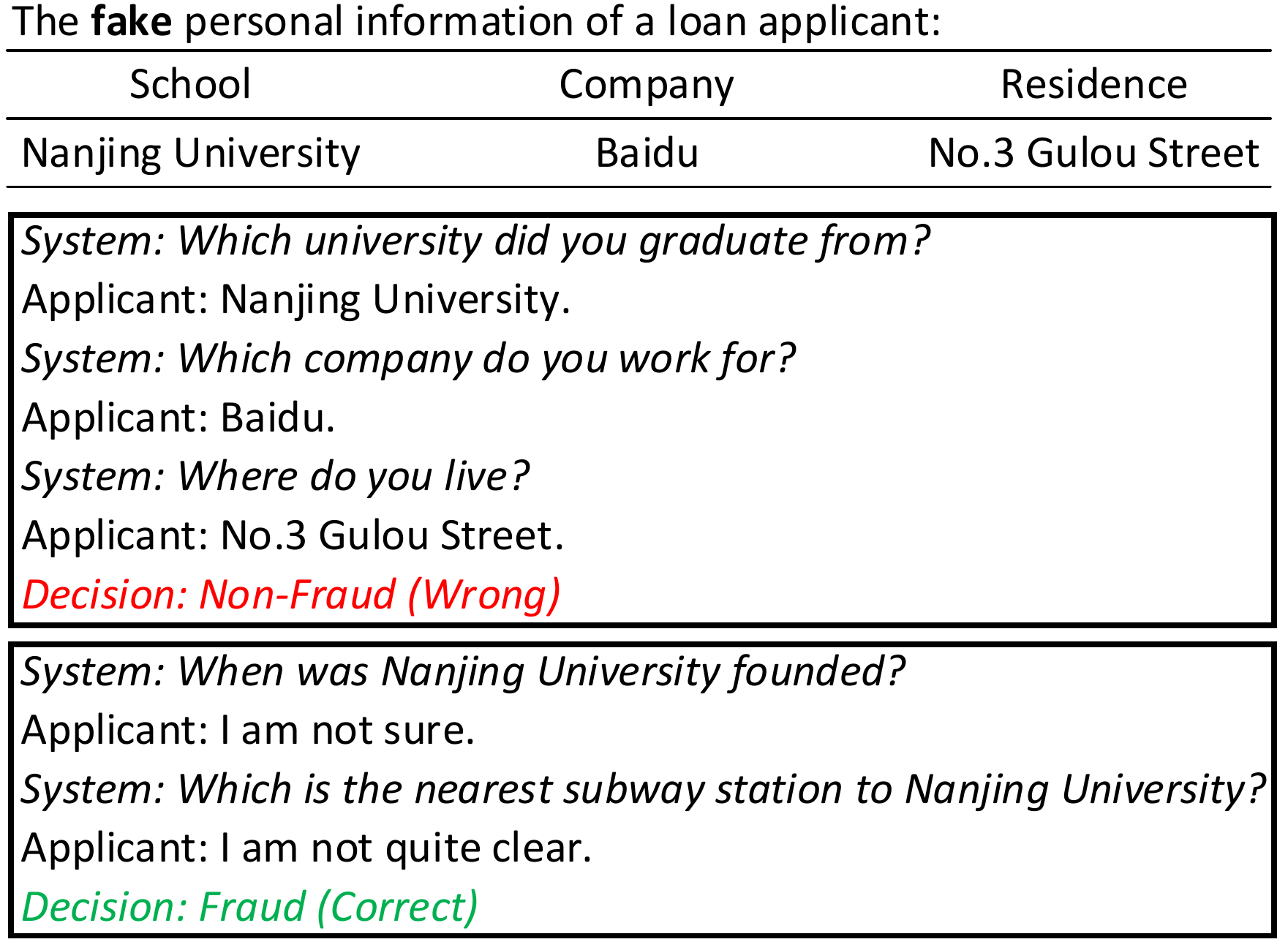}
	\caption{Dialogue examples of two possible fraud detection methods. The first one is directly asking applicants about their personal information. The second one is asking applicants about questions that are related to their personal information.}
	\label{fig:introduce}
\end{figure}

Intuitively, a simple way to detect identity fraud in loan applications is directly asking applicants about their personal information. However, as shown in Fig.~\ref{fig:introduce}, this method is prone to errors because fraudsters may well know the fake information. Fortunately, we find fraudsters generally are not clear about answers to questions that are related to the fake information.\footnote{This finding is based on the premise that loan applicants answer questions without any help~(e.g., using automatic QA systems or information retrieval tools). In fact, this premise is reasonable in many real scenarios, such as dialogue with video calls and phone calls in which we can monitor the applicants with a camera or require them to answer questions within few seconds (e.g., 5 seconds). } We refer to these questions as \emph{derived questions}, which can be constructed based on triplets where the head entity is the personal information entity. For example, the first derived question about ``Nanjing University'' is based on (Nanjing University, FoundedDate, 1902). In Fig.~\ref{fig:introduce}, the applicant claims to graduate from ``Nanjing University'' but can not answer derived questions about this school. This fact indicates that the applicant is likely to be a fraudster.

Based on the above finding, we aim to design a dialogue system to detect identity fraud by asking derived questions. However, there are three major challenges in achieving this goal.

First, designing derived questions requires a high-quality KG. However, owing to the sparseness problem~\cite{ji2016knowledge,trouillon2017knowledge} of the KG, many entities have no triplets for derived question generation. Second, randomly selecting triplets to generate questions is feasible but it is not the optimal questioning strategies to detect fraudsters. Third, because of privacy issues, evaluating anti-fraud systems with real applicants is not practical. And existing user simulation methods~\cite{li2016user,georgila2006user,pietquin2006probabilistic} do not apply to our task. Hence, how to evaluate our systems efficiently is a problem.

To address the above problems, we first complete an existing KG with geographic information in an electronic map (Section~\ref{KG_Constructor}). In the new KG, nearly all personal information entities can find triplets for derived question generation. Then, based on the KG, we present structured dialogue management (Section~\ref{system_design}) to explore the optimal dialogue strategy with reinforcement learning. Specifically, our dialogue management consists of (1) the \emph{KG-based dialogue state tracker}~(KG-DST) that treats embeddings of nodes in the KG as dialogue states and (2) the \emph{hierarchical dialogue policy}~(HDP) where high-level and low-level agents unfold the dialogue together. Finally, based on intuitive analysis, we find the applicants' behavior is related to some factors (Section~\ref{Human_Experiments}). Thus, we introduce hypotheses to formalize the effect of these factors on the applicants' behavior and propose a heuristic user simulator to evaluate our systems.

Experiments have shown that the data-driven system significantly outperforms rule-based systems in the fraud detection task. Besides, the ablation study proves that the proposed dialogue management can improve the recognition accuracy and learning efficiency because of its ability to model structured information. We also analyze the behavior of our system and find the learned anti-fraud policy is interpretable and flexible.

To summarise, our main contributions are three-fold: (1) As far as we know, this is the first work to detect identity fraud through dialogue interactions. (2) We point out three major challenges of identity fraud detection and propose corresponding solutions. (3) Experiments have shown that our approach can detect identity fraud effectively.

\section{Knowledge Graph Constructor}
\label{KG_Constructor}
There are four types of personal information in a Chinese loan application form: ``School'', ``Company'', ``Residence'' and ``BirthPlace''. To generate derived questions, we link all personal information entities to nodes in an existing Chinese KG\footnote{\url{https://www.ownthink.com}} and crawl triplets that are directly related to them. However, owing to the fact that the KG is largely sparse, nearly a half of entities\footnote{Most of them are ``Residence'' and ``BirthPlace''.} cannot be linked. Thus we use wealthy geographic information about organizations and locations in electronic maps~(e.g., Amap\footnote{\url{https://www.amap.com}}) to complete the KG.

\begin{figure}[htbp]
	\centering
	\setlength{\abovecaptionskip}{0.2cm}
	\includegraphics[width=0.37\textwidth]{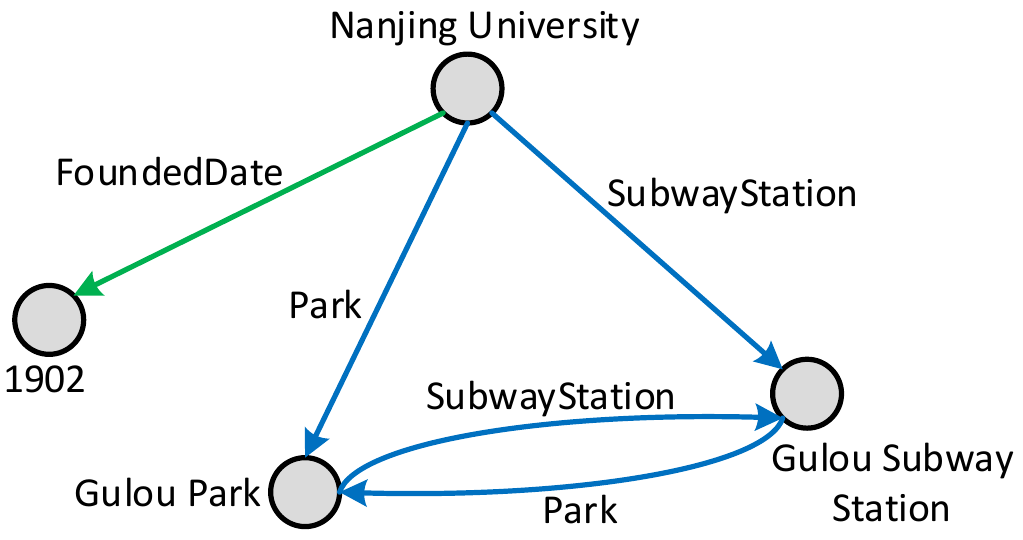}
	\caption{An example of the KG for Nanjing University. The green edge represents the triplet crawled from the existing KG and the blue edges represent the triplets generated based on a navigation electronic map.}
	\label{fig:kg_example}
\end{figure}
\begin{figure*}[htbp]
	\centering
	\setlength{\abovecaptionskip}{0.2cm}
	\includegraphics[width=0.95\textwidth]{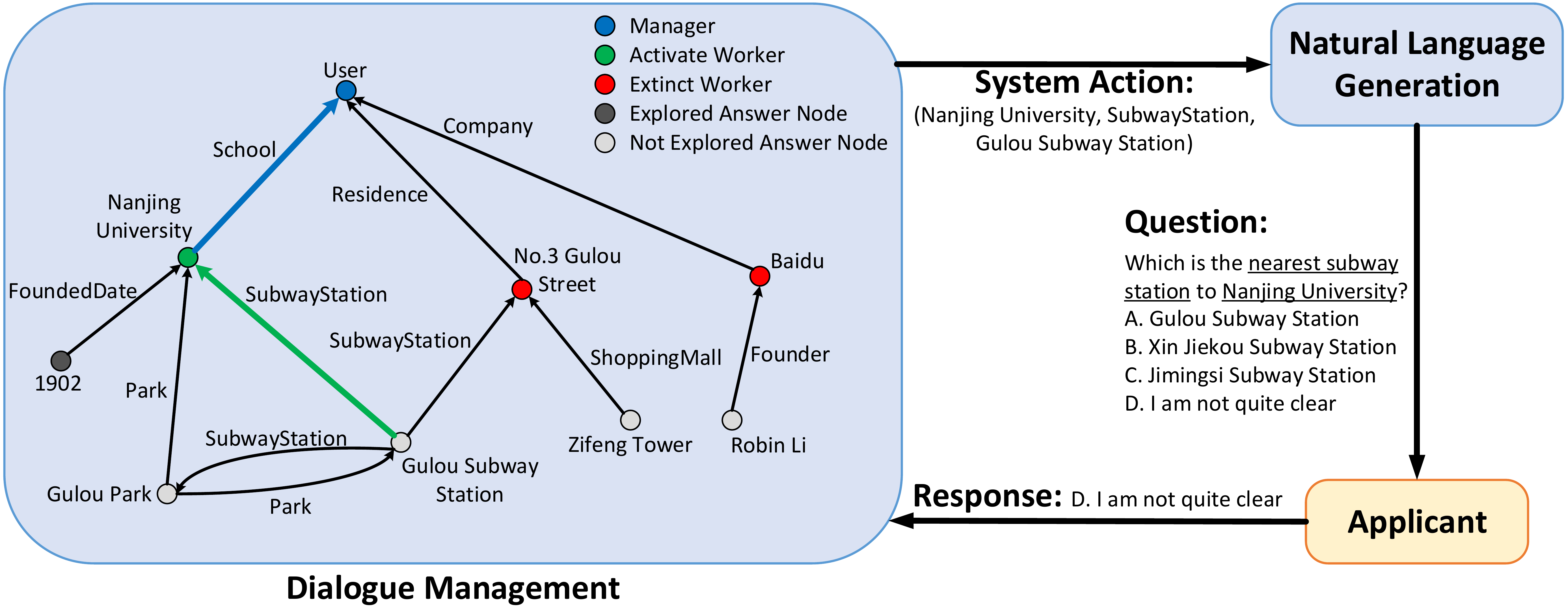}
	\caption{Overview of our approach. To build the directed graph for dialogue management, we reverse directions of all edges in the original personal KG to make the head entity read information from its tail entities. Besides, we add a special node ``User'' and new edges to represent the applicant's personal information. In this graph, the direction of each edge is the direction of message passing in KG-DST. The blue and green edges indicate that two agents select nodes to unfold the dialogue according to HDP.}
	\label{fig:model}
\end{figure*}

Specifically, for each personal information entity, we first crawl its points of interest~(POI\footnote{The POI are the specific locations~(e.g., subway stations) that someone may find useful in navigation systems.}) within one kilometer and the POI types in the Amap. If there are multiple POI for the same type, we only keep the nearest one. Then we generate triplets in the form of (\$Personal Information Entity\$, \$POI type\$, \$POI\$) to indicate the fact that the nearest \$POI type\$ to the \$Personal Information Entity\$ is \$POI\$. Besides, for any two entities, if the distance between them is less than 100 meters, we generate two triplets to represent the bi-directional adjacency relation between them. In the end, as shown in Fig.~\ref{fig:kg_example}, we combine triplets from the two information sources~(the Chinese KG and the electronic map) to construct a new KG. In this KG, nearly all personal information entities can be linked. And for each relation\footnote{After the data cleaning, there are 40 relations in all.}, we design a language template for the question generation.

\section{Dialogue System Design}
\label{system_design}
The overview of our system is shown in Fig.~\ref{fig:model}. The core of the system is dialogue management which is organized as a directed graph $G(\mathbb{V}, \mathbb{E})$. In each turn, our system first infers dialogue states with the \emph{KG-based dialogue state tracker} by computing embeddings of nodes. In this graph, the embedding of ``User'' node is the dialogue state of a high-level agent~(manager), and the embeddings of nodes adjacent to ``User''~(named as personal information nodes) are the dialogue states of low-level agents~(workers). Then our system unfolds the dialogue according to the \emph{hierarchical dialogue policy}. Concretely, the manager first selects a personal information node~(e.g., ``Nanjing University'') as the worker, and then the worker will select a node~(e.g., ``Gulou Subway Station'') from its predecessors~(named as answer nodes). After that, the sampled nodes of two agents form the final system action~(a triplet). Next, based on the triplet and a predefined template, the natural language generation module will give a multiple-choice question to the applicant. After the applicant gives a response, the embeddings of all nodes will be updated to generate new dialogue states for the next turn.

\subsection{KG-based Dialogue State Tracker}
There are three types of nodes in $G(\mathbb{V}, \mathbb{E})$: the ``User'' node $v_u$, the personal information node $v_p\in \mathbb{V}_p$ and the answer node $v_a\in \mathbb{V}_a$. In the $t\text{-th}$ turn, KG-DST first gives an initial embedding to $v\in \mathbb{V}_p\cup\mathbb{V}_a$. The initial embedding is the concatenation of \emph{static features} and \emph{dialogue features}. Then, $v$ will gather information from its predecessors $N(v)$. After multiple \emph{message passings}, we get its final embedding $E_t(v)$. Next, $v_u$ will aggregate information from $\mathbb{V}_p$ to generate its embedding $E_t(v_u)$. Finally, $E_t(v_u)$ and $E_t(v_p)$ are the dialogue states of the manager and worker respectively.

\begin{description}[leftmargin=0pt,listparindent=\parindent,parsep=0pt]
	\item [Static Features.] Specifically, for $v\in \mathbb{V}_p\cup\mathbb{V}_a$, the static features include the degree and type. Besides, for $v_a\in \mathbb{V}_a$, we use the ``spread degree on the internet'' to distinguish different answer nodes because we find there is an obvious correlation between this ``spread degree'' feature and applicants' behavior in our human experiments (Section~\ref{Human_Experiments}). To get the ``spread degree'' feature, we first treat the answer node $v_a$ and its adjacent personal information node $v_p$ as the keyword\footnote{In fact, the keyword is the head entity and tail entity of a triplet. For example, for the answer node ``1902'', the keyword is ``Nanjing University 1902''.}, and then search it in the search engine. The number\footnote{If there are multiple keywords for an answer node~(e.g., ``Gulou Subway Station''), we take the average.} of the retrieved results will be the ``spread degree'' feature of $v_a$. In the end, each static feature is encoded as a one-hot vector and they are concatenated to form a vector $S_t(v)$.
	\item [Dialogue Features.] The dialogue features record the dynamic information of $v\in \mathbb{V}_p\cup\mathbb{V}_a$ during the dialogue. Specifically, dialogue features include whether the node has been explored by the manager or workers and whether the node appeared in the system action of the last turn. In addition, for $v_p\in \mathbb{V}_p$, the dialogue features include the interaction turns of the corresponding worker and the number of correctly/incorrectly answered questions about $v_p$. For $v_a\in \mathbb{V}_a$, the dialogue features include whether applicants know $v_a$ is the answer to a derived question. Similarly, dialogue features will be encoded as a one-hot vector $D_t(v)$.
	\item [Message Passing.] In Fig.~\ref{fig:model}, the applicant does not know ``Gulou Subway Station'' is the nearest subway station to ``Nanjing University''. In such case, the personal information about ``School'' may be fake. Besides, for another question ``What's the nearest park to Nanjing University?'', the applicant may not know the answer because the distance between ``Gulou Park'' and ``Gulou Subway Station'' is less than 100 meters. Thus, we want the known information of ``Gulou Subway Station'' to be sent to its successors.
	
	Specifically, for $v\in \mathbb{V}_p\cup\mathbb{V}_a$, we compute its embedding recursively as follows:
	\begin{equation}
		\small
		E_t^k(v) = \max \limits_{v'\in N(v)}\,\tanh \left(\mathbf{W}^kE_t^{k-1}(v')\right)
	\end{equation}
	where $E_t^k(v)$ is the depth-$k$ node embedding in the $t\text{-th}$ turn, $N(v)$ denotes the set of nodes adjacent to $v$, $\mathbf{W}^k$ is the parameter in the $k\text{-th}$ iteration and the aggregate function is the element-wise max operation. The final node embedding is the concatenation of embeddings at each depth:
	\begin{equation}
		\small
		\label{eq:2}
		E_t(v) = \left[E_t^0(v),...,E_t^K(v)\right]
	\end{equation}
	where $E_t^0(v)=\left[S_t(v),D_t(v)\right]$ and $K$ is a hyperparameter.
	
	After getting the embedding of $v_p\in \mathbb{V}_p$, we compute the embedding of $v_u$ by aggregating information from $\mathbb{V}_p$:
	\begin{equation}
		\small
		E_t(v_u) = \max \limits_{v_p\in \mathbb{V}_p}\,\tanh \left(\mathbf{W}^pE_t(v_p)\right)
	\end{equation}
	where $\mathbf{W}^p$ is the parameter. 
	
	In the end, $E_t(v_p)$ is the worker's dialogue state which contains information of a part of the graph and $E_t(v_u)$ is the manager's dialogue state which contains information of the whole graph.
\end{description}

\subsection{Hierarchical Dialogue Policy}
After obtaining the dialogue states and node embeddings, our system will unfold the dialogue according to a hierarchical policy.

Specifically, the manager first selects $v_p\in \mathbb{V}_p$ as a worker to verify the identity state of $v_p$ according to a high-level policy $\pi^m$. Then, the worker will choose some answer nodes from its predecessors $N(v_p)$ to generate questions about $v_p$ according to a low-level policy $\pi^w$. If the worker gives the decision $d^w\in \{\text{Fraud}, \text{Non-Fraud}\}$ about the identity state of $v_p$, $\pi^w$ will end and the manager will select a new worker again or give the final decision. If the manager gives the final decision $d^m\in \{\text{Fraud}, \text{Non-Fraud}\}$ about the applicant's identity state, $\pi^m$ will end. Formally, $\pi^m$ and $\pi^w$ are defined as follows:
\begin{equation}
	\small
	\begin{split}
		\pi^m_t(v_p|E_t(v_u))&\propto \exp{\left(\mathbf{W}^m\left[E_t(v_u),E_t(v_p)\right]+b^m\right)} \\
		\pi^m_t(d^m|E_t(v_u))&\propto \exp{\left(\mathbf{W}^m\left[E_t(v_u),E(d^m)\right]+b^m\right)} \\
		\pi^w_t(v_a|E_t(v_p))&\propto \exp{\left(\mathbf{W}^w\left[E_t(v_p),E_t(v_a)\right]+b^w\right)} \\
		\pi^w_t(d^w|E_t(v_p))&\propto \exp{\left(\mathbf{W}^w\left[E_t(v_p),E(d^w)\right]+b^w\right)}
	\end{split}
\end{equation}
where $\{\mathbf{W}^m,\mathbf{W}^w,b^m,b^w,E(d^m),E(d^w)\}$ are parameters, $E_t(v_u)$ and $E_t(v_p)$ are dialogue states of the manager and worker in the $t\text{-th}$ turn, $E(d^m)$ is the encoding of the manager's terminal action which has the same dimension as $E_t(v_p)$, and $E(d^w)$ is the encoding of the worker's terminal action which has the same dimension as $E_t(v_a)$.

Besides, to prevent the two agents from making decisions in haste, domain rules are applied to their dialogue policies by ``Action Mask''~\cite{williams2017hybrid}. Specifically, domain rules are defined as follows. First, only after all or at least three answer nodes related to a worker have been explored can the worker make the decision. Second, only after all workers have made decisions or at least one worker's decision is ``Fraud'' can the manager make the final decision.

\section{Training}
\subsection{Reward Function}
We expect the system can give correct decisions about applicants within minimum turns. Thus, at the end of each dialogue, the manager receives a positive reward $r_{crt}^m$ for correct decision, or a negative reward $-r_{wrg}^m$ for wrong decision. If the manager selects a worker to unfold the dialogue in the $t\text{-th}$ turn and the worker gives $n_t^w$ questions to the applicant, the manager will receive a negative reward $-n_t^w*r_{turn}$. Besides, we provide an internal reward to optimize the low-level policy. Specifically, if the worker gives a correct decision about the corresponding personal information, it will receive a positive reward $r_{crt}^w$. Otherwise, it will receive a negative reward $-r_{wrg}^w$. And in each turn, the worker receives a negative reward $-r_{turn}$ to encourage shorter interactions.

\subsection{Reinforcement Learning}
The two agents can be trained with policy gradient~\cite{williams1992simple} approach as follows:
\begin{equation}
	\small
	\begin{split}
		\nabla\pi^m_t&=\Big(R_t^m-V_t^m\big(E_t(v_u)\big)\Big)\nabla\log\pi^m_t\big(a_t^m|E_t(v_u)\big) \\
		\nabla\pi^w_t&=\Big(R_t^w-V_t^w\big(E_t(v_p)\big)\Big)\nabla\log\pi^w_t\big(a_t^w|E_t(v_p)\big)
	\end{split}
\end{equation}
where $R_t^m$ and $R_t^w$ are the discounted returns of two agents, $a_t^m$ and $a_t^w$ are their sampled actions, $V_t^m\big(E_t(v_u)\big)$ and $V_t^w\big(E_t(v_p)\big)$ are value networks which are optimized by minimizing mean-square errors to $R_t^m$ and $R_t^w$ respectively.

\subsection{Pre-Training}
\label{Training}
Before reinforcement learning~(RL), supervised learning~(SL) is applied to mimic dialogues provided by a rule-based system. Rules are defined as follows. First, the manager selects a worker randomly. Then, the worker will select answer nodes randomly to generate questions. Let $n_{crt}$/$n_{wrg}$ denotes the number of correctly/incorrectly answered questions in this worker's decision process. If $|n_{crt} - n_{wrg}|\geq3$ or all answer nodes related to this worker have been explored, the worker will give its decision. If $n_{crt}<n_{wrg}$, the worker's decision will be ``Fraud'' and the manager's decision will be ``Fraud'' too. Otherwise, the worker's decision will be ``Non-Fraud'' and the manager will choose a new worker to continue the dialogue. In the end, if all workers' decisions are both ``Non-Fraud'', the manager's decision will be ``Non-Fraud''.

\section{Experiments and Results}
\subsection{User Simulator and Human Experiments}
\label{Human_Experiments}
Simulating users' behavior is an efficient way to evaluate dialogue systems. In our task, the applicants' behavior is answering derived questions. Thus, the key of user simulator is to estimate the probability $p(\boldsymbol{k}_i)$, where $\boldsymbol{k}_i$ is a binary random variable which denotes whether or not the applicant knows the triplet fact $\boldsymbol{t}_i$ behind a question $\boldsymbol{q}_i$.

Intuitively, $p(\boldsymbol{k}_i)$ depends on three factors. First, if the applicant's identity state is ``Non-Fraud'', $p(\boldsymbol{k}_i=1)$ will be greater than $p(\boldsymbol{k}_i=0)$. Second, the wider a triplet fact $\boldsymbol{t}_i$ spreads on the internet, the more likely applicants know it. For example, almost all of applicants know (Baidu, Founder, Robin Li) because there are a lot of web pages containing this fact on the internet. Third, if applicants know other triplets that are related to $\boldsymbol{t}_i$, they may well know $\boldsymbol{t}_i$ because it is easy to deduce $\boldsymbol{t}_i$ based on what they know. For example, if applicants know (Nanjing University, Park, Gulou Park) and (Gulou Park, SubwayStation, Gulou Subway Station),  they may well know (Nanjing University, SubwayStation, Gulou Subway Station).

To formalize the effect of the three factors on applicants' behavior, we introduce three hypotheses: (1) For both fraudsters and normal applicants, $p(\boldsymbol{k}_i=1)$ is proportional to the ``spread degree'' of $\boldsymbol{t}_i$. (2) The ``spread degree'' of $\boldsymbol{t}_i$ can be approximated by the number of retrieved results~(denoted as $\text{Freq}(e^h_i,e^t_i)$) in search engine where the keyword is the head entity $e^h_i$ and the tail entity $e^t_i$ of $\boldsymbol{t}_i$. (3) For any three triplets, if they form a closed loop~(regardless of directions) and applicants know two of them, the applicants must know all of them.

\begin{figure}[!h]
	\centering
	\setlength{\abovecaptionskip}{0.2cm}
	\includegraphics[width=0.48\textwidth]{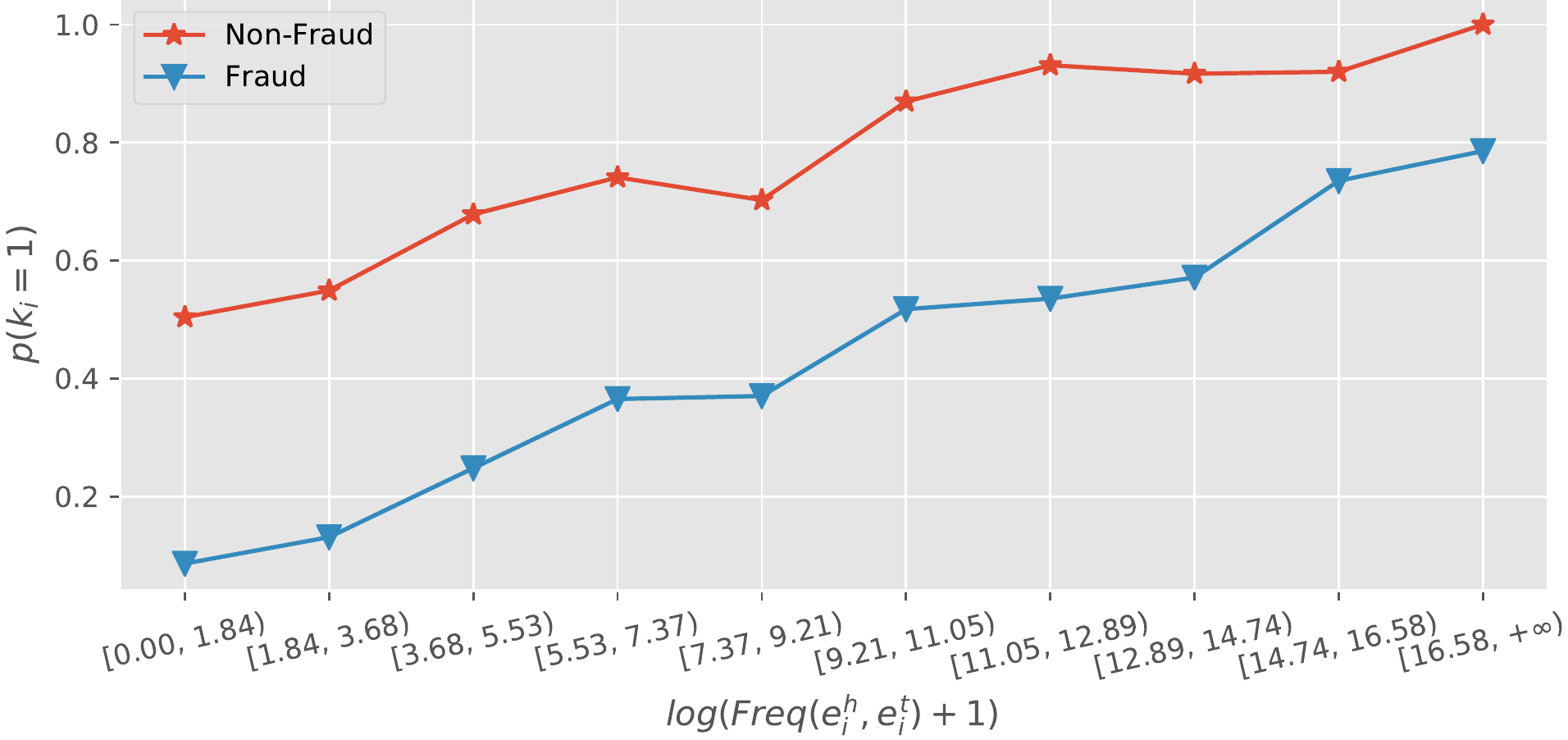}
	\caption{The relations between $\text{Freq}(e^h_i,e^t_i)$ and $p(\boldsymbol{k}_i)$ for two kinds of applicants. $\text{Freq}(e^h_i,e^t_i)$ is used to approximate the ``spread degree'' of $\boldsymbol{t}_i$. $p(\boldsymbol{k}_i=1)$ indicates the probability that applicants know $\boldsymbol{t}_i$.}
	\label{fig:human_experiments}
\end{figure}

To generate simulated loan applicants, we first estimate the function relations between $p(\boldsymbol{k}_i=1)$ and $\text{Freq}(e^h_i,e^t_i)$ via human experiments. Specifically, we ask 31 volunteers to answer derived questions\footnote{There are 1516 derived questions in all.} about their own and others' personal information. And then, for the question $\boldsymbol{q}_i$, we place it into a discrete bin according to the logarithm of $\text{Freq}(e^h_i,e^t_i)$. In each bin, we use the ratio of correctly answered questions to approximate $p(\boldsymbol{k}_i=1)$. In the end, the relations are shown in Fig.~\ref{fig:human_experiments}. We can find that the statistical distributions of real behavior patterns of normal applicants and fraudsters are distinguishable and the results agree with our first two intuitions.
\begin{figure*}[!t]
	\centering
	\subfigure{
		\label{fig:accuracy}
		\includegraphics[width=0.43\textwidth]{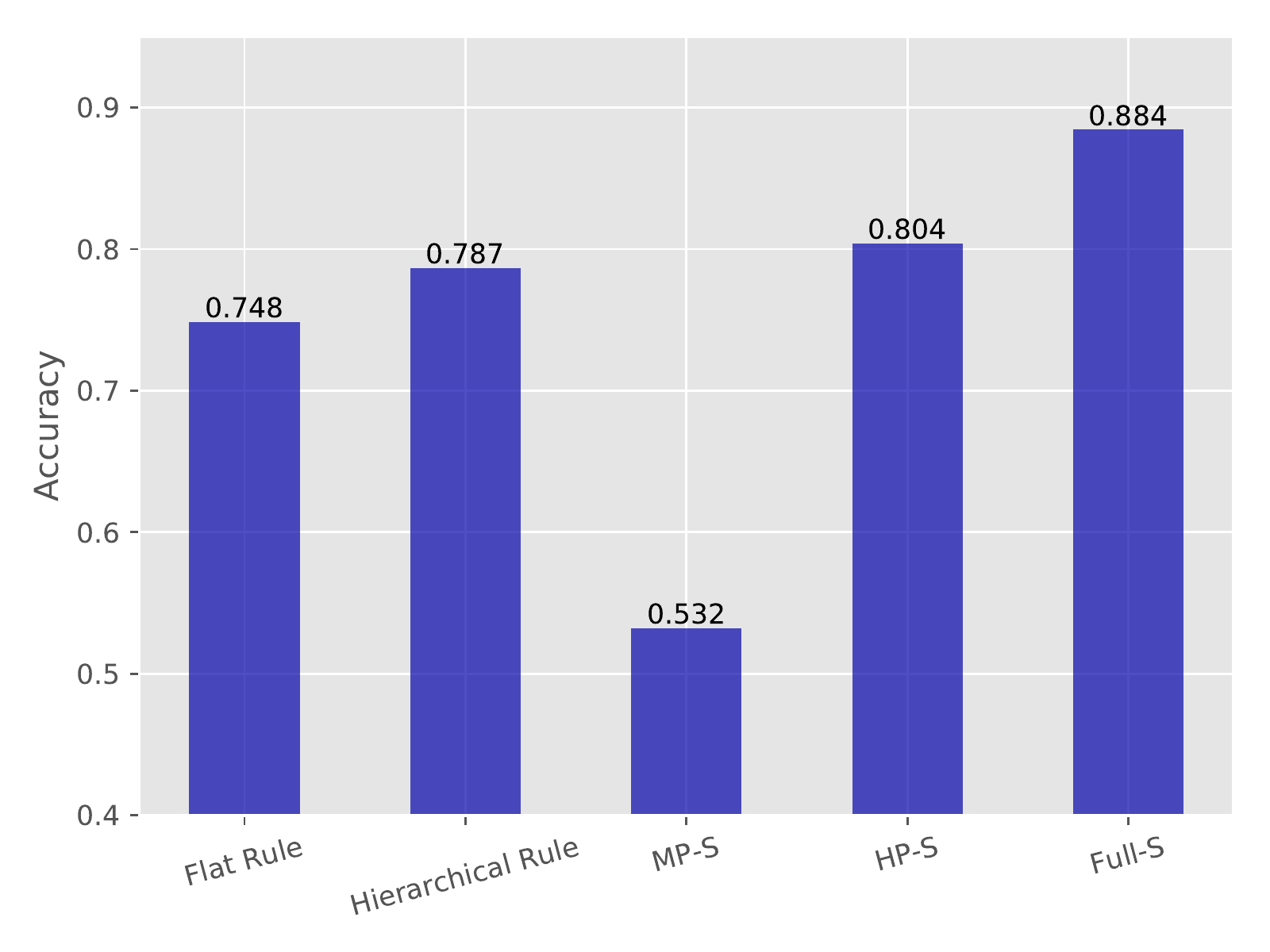}}
	\subfigure{
		\label{fig:turns}
		\includegraphics[width=0.43\textwidth]{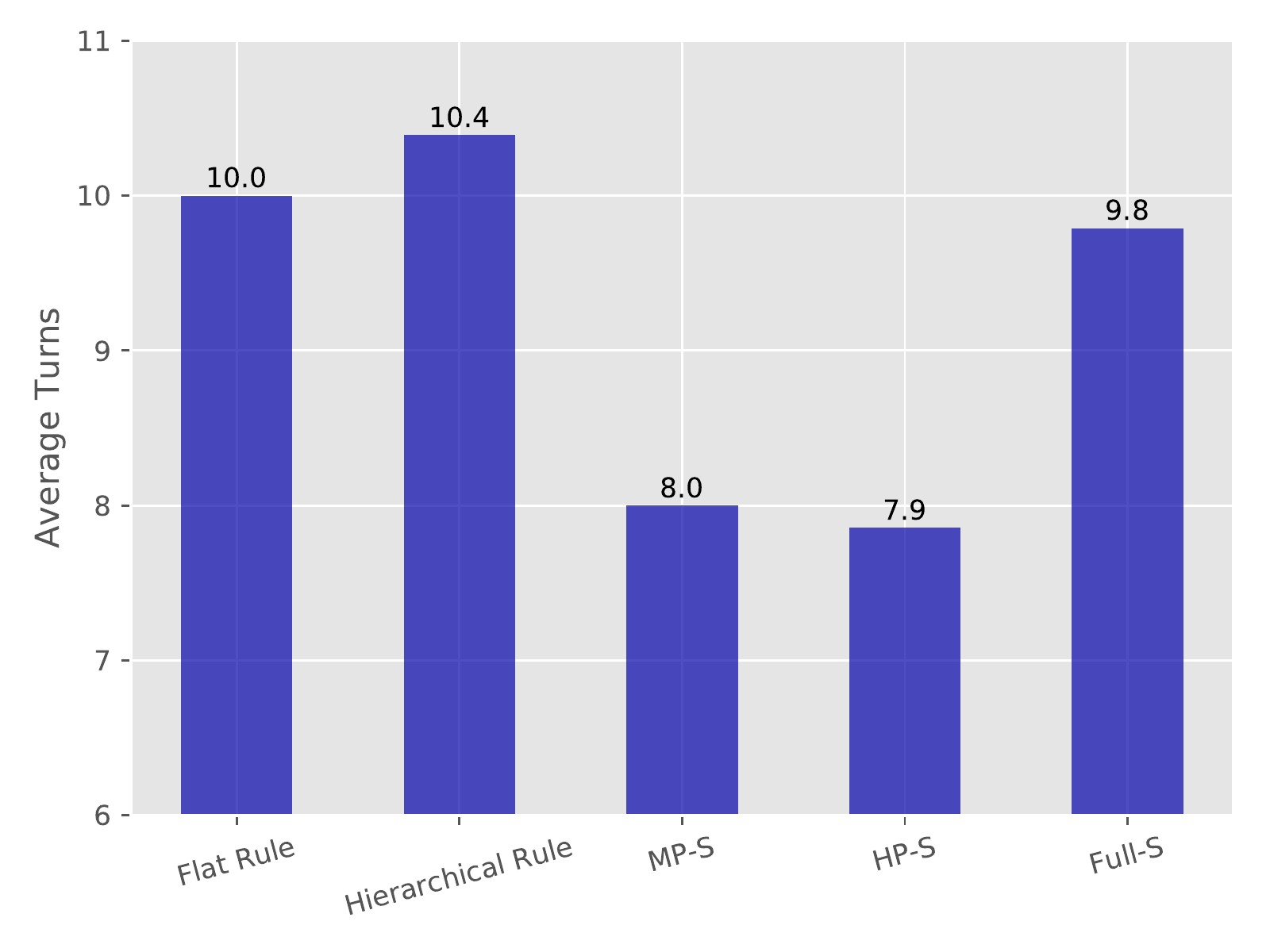}}
	\caption{Performance of different systems. Tested on 10 epochs using the best model during training.}
	\label{fig:main_results}
\end{figure*}

Then, we get simulated loan applicants\footnote{Note that we can generate any number of simulated applicants based on one applicant's personal information.} following a ``\textit{sampling and calibration}'' manner. Specifically, given an applicant's personal information, we first sample the identity state randomly. If the sampling result is ``Fraud'', we will sample $1\sim4$ information item(s) randomly to be the fake information. Generally, forging information about ``School'' and ``Company'' may result in a larger loan. Thus, when sampling the fake information, the sampling probability of ``School'' and ``Company'' is twice the sampling probability of ``Residence'' and ``BirthPlace''. Then, for each personal information and its related triplet $\boldsymbol{t_i}$, we sample $\boldsymbol{k_i}$ based on (1) whether the personal information is fake (2) $\text{Freq}(e^h_i,e^t_i)$ and (3) the corresponding function relation in Fig.~\ref{fig:human_experiments}. Because the sampling results $\{\boldsymbol{k_1},...,\boldsymbol{k_n}\}$ are independent from each other, there may be the situations where the sampling results do not satisfy the rule defined in our third hypothesis. If that happens, we calibrate it until all sampling results agree with the hypothesis. Finally, if $\boldsymbol{k}_i=1$, the applicant will give the correct answer to the question $\boldsymbol{q}_i$. Otherwise, the applicant's response is ``D. I am not quite clear.''.

\subsection{Baselines}
We compare our model~(denoted as Full-S) with two rule-based baselines. In addition, to study the effect of message passing and hierarchical policy on the model training, we compare Full-S with two neural baselines for the ablation study.
\begin{itemize}[parsep=0pt,itemsep=0pt,topsep=0pt]
	\item {Flat Rule}: The system selects 10 questions randomly to ask applicants. If the number of correctly answered questions is fewer than the number of incorrectly answered questions, the system's decision will be ``Fraud''. Otherwise, the system's decision will be ``Non-Fraud''.
	\item {Hierarchical Rule}: A rule-based system which uses a hand-crafted hierarchical policy to unfold dialogues. As shown in Section~\ref{Training}, we use this system to pre-train Full-S.
	\item {MP-S}: A neural dialogue system which uses message passing to infer dialogue states but uses a flat policy to unfold dialogues. That is, the manager selects answer nodes directly to generate derived questions.
	\item {HP-S}: A neural dialogue system which uses the hierarchical policy to unfold dialogues but does not use message passing to infer dialogue states. That is, $K$ is 0 in Eq.~\ref{eq:2}.
\end{itemize}

\subsection{Implementation Details}
We collect 906 applicants' personal information, and randomly select 706 for training, 100 for dev, and 100 for test. In each batch, we sample 32 applicants' information for simulation. The maximum interaction turns of the system and the worker are 40 and 10 respectively. The iteration depth $K$ is 2 in message passing. In the reward function, $r_{crt}^m=3$, $r_{wrg}^m=3$, $r_{crt}^w=1$, $r_{wrg}^w=1$, $r_{turn}=0.1$. The discount factors are 0.999 and 0.99 for the manager and worker respectively. All neural dialogue systems are both pre-trained with rule-based systems for 20 epochs. We pre-train MP-S with Flat Rule because they both use the flat policy. Besides, we pre-train HP-S and Full-S with Hierarchical Rule because they both use the hierarchical policy. In the RL stage, all neural dialogue systems are trained for 300 epochs. When testing, we repeat 10 epochs and take the average.

\subsection{Test Performance}
We compare Full-S with baselines in terms of two metrics: recognition accuracy and average turns.

Fig.~\ref{fig:main_results} shows the test performance. We can see that the accuracy of Flat Rule is lower than Hierarchical Rule, and the accuracy of the data-driven counterpart of Flat Rule~(MP-S) is just slightly higher than randomly guessing. It means that using the hierarchical policy to unfold dialogues is necessary for our task. Besides, HP-S achieves a higher accuracy than its rule-based counterpart~(Hierarchical Rule) within much fewer turns. It proves that the data-driven system is more efficient than the rule-based system. Finally, equipped with message passing and hierarchical policy, Full-S achieves the best accuracy. And it is interesting to note that Full-S requires more turns but achieves much higher accuracy than HP-S. One possbile reason is that HP-S may easily trap in local optimum without message passing to infer dialogue states.

\begin{figure}[!t]
	\centering
	\includegraphics[width=0.42\textwidth]{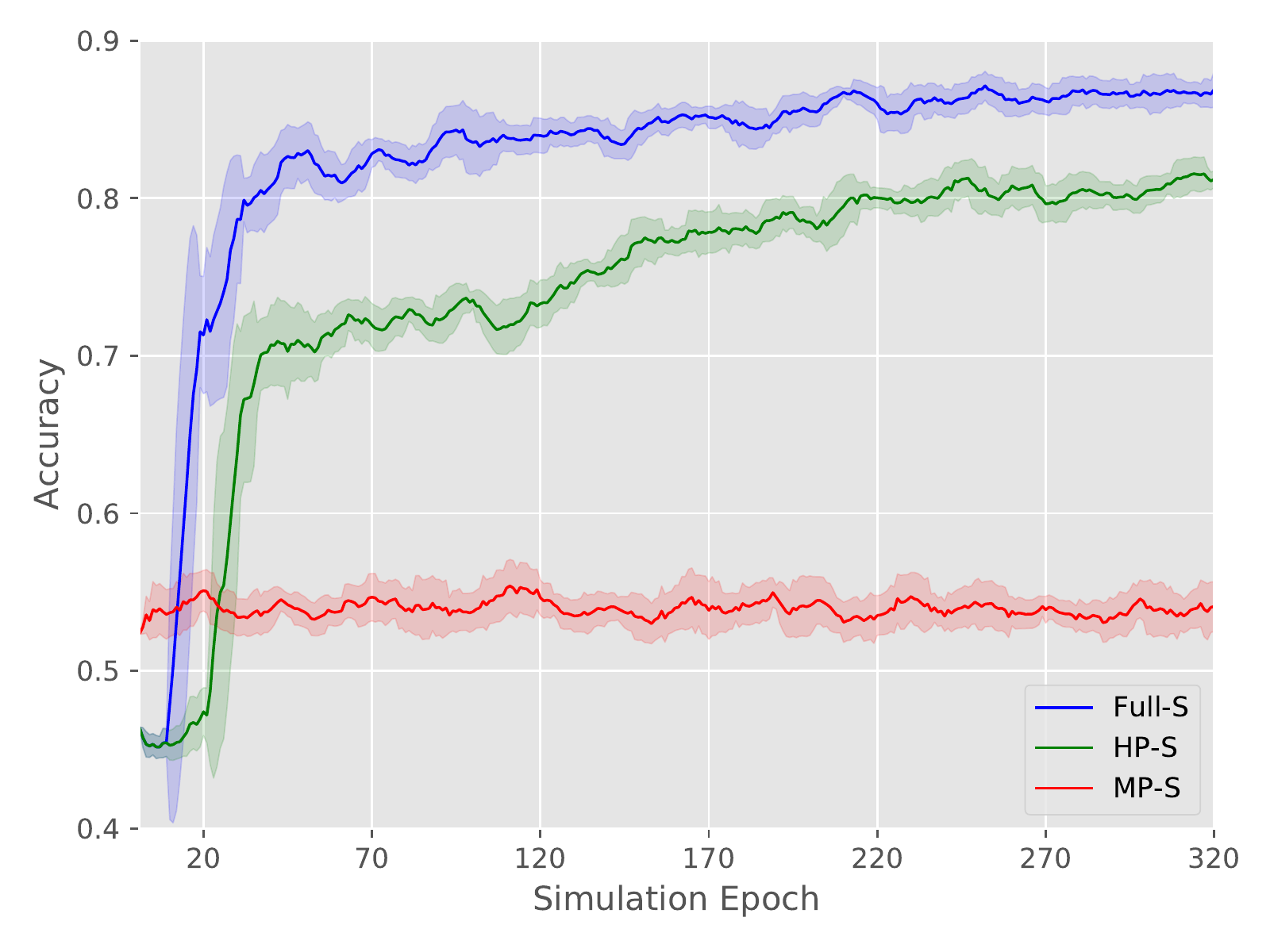}
	\caption{Accuracy curves of different neural models in dev set. The first ``20 epochs'' indicates the pre-training stage. The last ``300 epochs'' indicates the RL stage.}
	\label{fig:accuracy_curve}
\end{figure}

\subsection{Ablation Study}
To study the effect of message passing and hierarchical policy, we show the learning curves of three neural dialogue systems in Fig.~\ref{fig:accuracy_curve}. Each learning curve is averaged on 10 epochs.

We find that, compared with Full-S and HP-S, MP-S is unable to learn any useful dialogue policy during training. There are two reasons for this. First, the action space of flat policy is too large, which results that MP-S suffers from the sparse reward and long horizon issues. Second, without explicitly modeling the logic relation between the manager and workers, MP-S is prone to errors. Besides, we can see that the convergence speed of Full-S is faster than HP-S in both the pre-training and the RL stages. This is because message passing can model structured information of the KG, and hence Full-S is more efficient in policy learning.

\subsection{Manager's Policy Analysis}
To better understand the high-level dialogue policy, we analyze the manager's behavior in Full-S.

\begin{figure}[!t]
	\centering
	\includegraphics[width=0.42\textwidth]{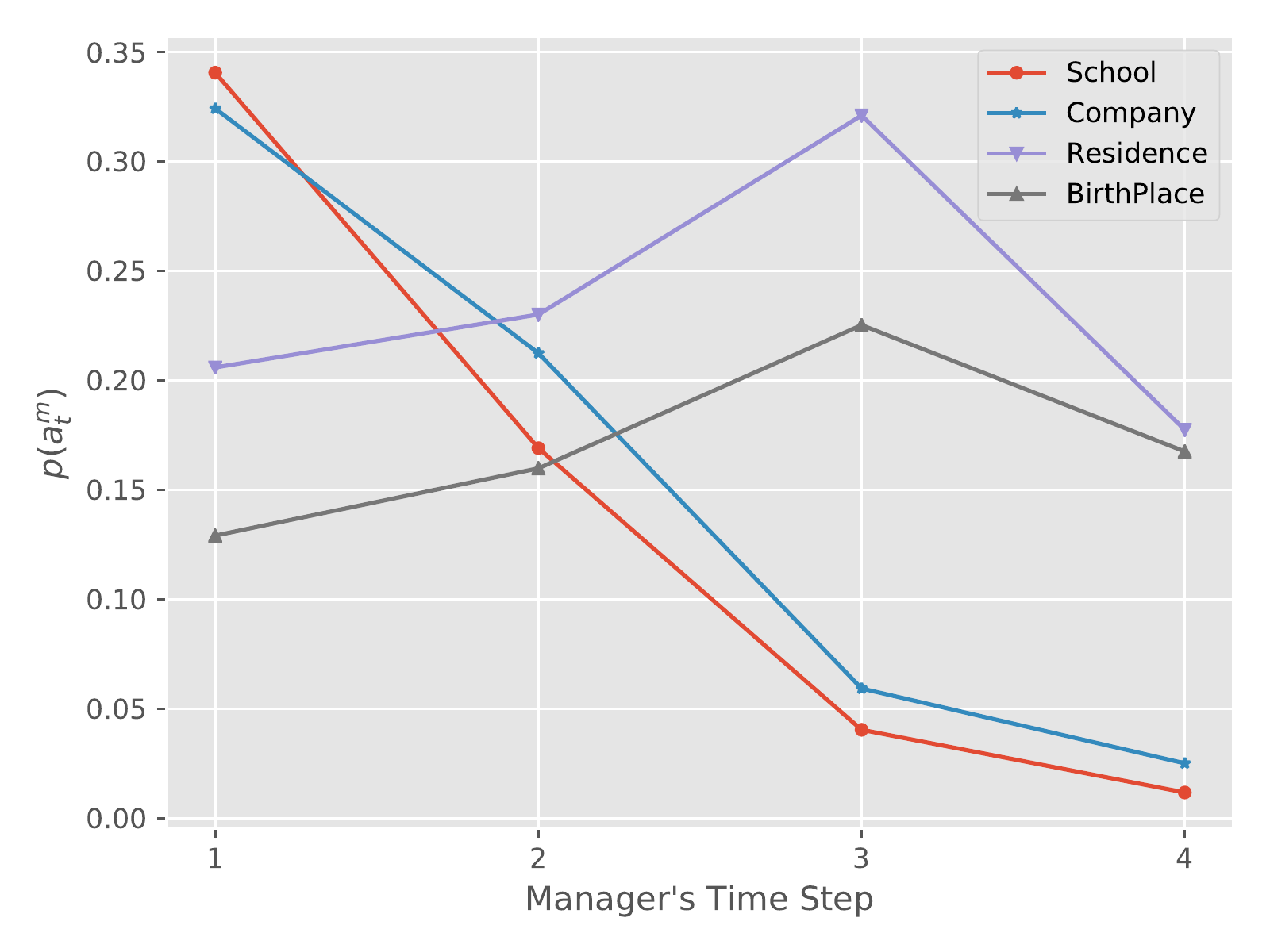}
	\caption{Manager's action probability curves. Each curve indicates the probability of selecting a piece of personal information to verify. For each curve, we take the average of all dialogues during testing.}
	\label{fig:manager_action_prob}
\end{figure}

First, we show the manager's action probability curves in Fig.~\ref{fig:manager_action_prob}. We can see that selecting ``School'' and ``Company'' to verify personal information has a priority over ``Residence'' and ``BirthPlace'' in the first decision step. And in the following two decision steps, the probabilities of selecting ``Residence'' and ``BirthPlace'' will increase. This is because simulated applicants tend to forge personal information about ``School'' and ``Company'' for a larger loan. Consequently, to discover fake information faster, the manager learns to prioritize different information items.

Second, intuitively the manager's policy should follow two logic rules in our task:
\begin{description}[leftmargin=0pt,parsep=0pt]
	\item[Rule1:] If a worker's decision is ``Fraud''~($Cond1$), the dialogue should end immediately and the manager's decision will be ``Fraud''~($RS1$).
	\item[Rule2:] If all workers' decisions are both ``Non-Fraud''~($Cond2$), the manager's decision will be ``Non-Fraud''~($RS2$).
\end{description}
To test whether the manager follows the two rules, we calculate the probabilities of $RS1$ and $RS2$ under $Cond1$ and $Cond2$ respectively. Specifically, in the test data, $p(RS1|Cond1)=0.95$ and $p(RS2|Cond2)=0.96$. It proves that the manager will adopt workers' suggestions in most situations.

\begin{table*}[!t]
	\centering
	\small
	\begin{tabular}{@{}ll@{}}
	\toprule
	\multicolumn{2}{c}{All triplets that are related to ``Shanghai Sports University'' (replaced with \$School\$ for short):} \\ \midrule
	(\$School\$, SuperMarket, Educational Supermarket)                 & (\$School\$, PetMarket, Seasons Garden)           \\
	(\$School\$, LocatedIn, Shanghai)                                  & (\$School\$, FoundedDate, 2002)                   \\
	(\$School\$, DigitalMall, JinLu Security)                          & (\$School\$, FruitShop, Xiao Liu Fruit)           \\
	(\$School\$, ConvenienceStore, HaoDe)                              & (Xiao Liu Fruit, ConvenienceStore, HaoDe)         \\
	(HaoDe, FruitShop, Xiao Liu Fruit)                                 &                                                   \\ \midrule
	\multicolumn{1}{c|}{HP-S}                                          & \multicolumn{1}{c}{ Full-S}                       \\ \midrule
	\multicolumn{1}{l|}{\emph{System: Which is the nearest pet market to \$School\$?}}     & \emph{System: Which is the nearest pet market to \$School\$?}    \\
	\multicolumn{1}{l|}{Applicant: I am not quite clear.} 		                           & Applicant: I am not quite clear.                                 \\
	\multicolumn{1}{l|}{\emph{System: Which is the nearest digital mall to \$School\$?}}   & \emph{System: Which is the nearest digital mall to \$School\$?}  \\
	\multicolumn{1}{l|}{Applicant: I am not quite clear.}                                  & Applicant: I am not quite clear.                                 \\
	\multicolumn{1}{l|}{\emph{System: Where is \$School\$ located?}}                       & \emph{System: Which is the nearest fruit shop to \$School\$?}    \\
	\multicolumn{1}{l|}{Applicant: Shanghai.}                                              & Applicant: Xiao Liu Fruit                                        \\
	\multicolumn{1}{l|}{}                                                                  &\emph{System: Which is the nearest supermarket to \$School\$?}    \\
	\multicolumn{1}{l|}{}                                                                  & Applicant: Educational Supermarket                               \\
	\multicolumn{1}{l|}{}                                                                  & \emph{System: When was \$School\$ founded?}                      \\
	\multicolumn{1}{l|}{}                                                                  & Applicant: 2002                                                  \\ \midrule
	\multicolumn{1}{c|}{Decison: Fraud~(Wrong)}                                            & \multicolumn{1}{c}{Decison: Non-Fraud~(Correct)}                 \\ \bottomrule
	\end{tabular}
	\caption{Examples of the low-level policies in two systems. Note that the information about ``School'' is not fake.}
	\label{tab:dialogue_example}
	\end{table*}

Meanwhile, we study cases where the manager does not follow the two rules and find some interesting phenomena. Specifically, if only one worker's decision is ``Fraud'' and the applicant can answer a few questions given by this worker, the manager's decision may be ``Non-Fraud''. Besides, if all workers' decisions are both ``Non-Fraud'' but the applicant can not answer most of the questions given by one worker, the manager's decision may still be ``Fraud''. In fact, when the two cases happen, the worker may make the wrong decision. However, the manager can still give the correct decision. It means the manager is robust to workers' mistakes.

\subsection{Worker's Policy Analysis}
To better understand the low-level dialogue policy and the effect of message passing on it, we compare workers' behaviors in HP-S and Full-S.

Table~\ref{tab:dialogue_example} shows an example of verifying personal information about ``School'' in HP-S and Full-S. We can see that the two systems give the same two questions in the first two turns. This is because the triplets behind the two questions are rarely known to fraudsters. It means that the low-level policies learn to give priority to such triplets for better distinguishing fraudsters from normal applicants. In the third turn, HP-S gives a question that is easy to answer for fraudsters and makes the wrong decision. However, Full-S notices the applicant gives the correct answer to a question that is hard to answer for fraudsters. Thus, Full-S does not make the decision in haste but continue the dialogue. Besides, it is worth noting that Full-S has not chosen (\$School\$, ConvenienceStore, HaoDe) to generate the derived question. This is because the message passing mechanism models the relation between ``HaoDe'' and ``Xiao Liu Fruit''. Specifically, because the two entities are closely related to each other, if applicants know ``Xiao Liu Fruit'', they may well know ``HaoDe''. Thus, there is no need to select this triplet anymore.

\section{Related work}
As far as we know, there is no published work about detecting identity fraud via interactions. We describe the two most related directions as follows:

\textbf{Deception Detection.} Detecting deception is a longstanding research goal in many artificial intelligence topics. Existing work has mainly focused on extracting useful features from non-verbal behaviors~\cite{meservy2005deception,lu2005blob,bhaskaran2011lie}, speech cues~\cite{levitan2018acoustic,graciarena2006combining} or both~\cite{krishnamurthy2018deep,perez2015verbal} to train a classification model. In their work, the definition of deception is telling a lie. Besides, existing work requires labeled data, which is often hard to get. In contrast, we focus on detecting identity fraud through multi-turn interactions and use reinforcement learning to explore the anti-fraud policy without any labeled data.

\textbf{Dialogue System.} Our work is also related to task-oriented dialogue systems~\cite{young2013pomdp,wen2017network,li2017end,gavsic2011line,wang-etal-2018-teacher,wang-etal-2019-incremental}. Existing systems have mainly focused on slot-filling tasks~(e.g., booking a hotel). In such tasks, a set of system actions can be pre-defined based on the business logic and slots. In contrast, the system actions in our task are selecting nodes in the KG to generate questions. Thus, the structured information is important in our task. Besides, some works also try to model structured information in dialogue systems. For example, \citeauthor{peng2017composite}~(\citeyear{peng2017composite}) used hierarchical reinforcement learning~\cite{vezhnevets2017feudal,kulkarni2016hierarchical,florensa2017stochastic} to design multi-domain dialogue management. \citeauthor{chen2018structured}~(\citeyear{chen2018structured}) used graph neural networks~\cite{battaglia2018relational,li2015gated,scarselli2009graph,niepert2016learning} to improve the sample-efficiency of reinforcement learning. \citeauthor{he-etal-2017-learning}~(\citeyear{he-etal-2017-learning}) used DynoNet to incorporate structured information in the collaborative dialogue setting. Compared with them, our method is a combination of the graph neural networks and hierarchical reinforcement learning, and experiments prove that they both work in the novel dialogue task.

\section{Conclusion}
This paper proposes to detect identity fraud automatically via dialogue interactions. To achieve this goal, we present structured dialogue management to explore anti-fraud dialogue strategies based on a KG with reinforcement learning and a heuristic user simulator to evaluate our systems. Experiments have shown that end-to-end systems outperform rule-based systems and the proposed dialogue management can learn interpretable and flexible dialogue strategies to detect identity fraud more efficiently. We believe that this work is a basic first step in this promising research direction and will help promote many real-world applications.

\section{Acknowledgments}
The research work described in this paper has been supported by the National Key Research and Development Program of China under Grant No. 2017YFB1002103. We would like to thank Shaonan Wang, Yang Zhao, Haitao Lin, Cong Ma, Lu Xiang and Junnan Zhu for their suggestions on this paper and Fenglv Lin for his help in the POI dataset construction.

\bibliography{emnlp-ijcnlp-2019}

\begin{thebibliography}{30}
\expandafter\ifx\csname natexlab\endcsname\relax\def\natexlab#1{#1}\fi

\bibitem[{Battaglia et~al.(2018)Battaglia, Hamrick, Bapst, Sanchez-Gonzalez,
  Zambaldi, Malinowski, Tacchetti, Raposo, Santoro, Faulkner
  et~al.}]{battaglia2018relational}
Peter~W Battaglia, Jessica~B Hamrick, Victor Bapst, Alvaro Sanchez-Gonzalez,
  Vinicius Zambaldi, Mateusz Malinowski, Andrea Tacchetti, David Raposo, Adam
  Santoro, Ryan Faulkner, et~al. 2018.
\newblock Relational inductive biases, deep learning, and graph networks.
\newblock \emph{arXiv preprint arXiv:1806.01261}.

\bibitem[{Bhaskaran et~al.(2011)Bhaskaran, Nwogu, Frank, and
  Govindaraju}]{bhaskaran2011lie}
Nisha Bhaskaran, Ifeoma Nwogu, Mark~G Frank, and Venu Govindaraju. 2011.
\newblock Lie to me: Deceit detection via online behavioral learning.
\newblock In \emph{Face and Gesture 2011}, pages 24--29. IEEE.

\bibitem[{Chen et~al.(2018)Chen, Tan, Long, and Yu}]{chen2018structured}
Lu~Chen, Bowen Tan, Sishan Long, and Kai Yu. 2018.
\newblock Structured dialogue policy with graph neural networks.
\newblock In \emph{Proceedings of the 27th International Conference on
  Computational Linguistics}, pages 1257--1268.

\bibitem[{Florensa et~al.(2017)Florensa, Duan, and
  Abbeel}]{florensa2017stochastic}
Carlos Florensa, Yan Duan, and Pieter Abbeel. 2017.
\newblock Stochastic neural networks for hierarchical reinforcement learning.
\newblock \emph{arXiv preprint arXiv:1704.03012}.

\bibitem[{Ga{\v{s}}i{\'c} et~al.(2011)Ga{\v{s}}i{\'c},
  Jur{\v{c}}{\'\i}{\v{c}}ek, Thomson, Yu, and Young}]{gavsic2011line}
Milica Ga{\v{s}}i{\'c}, Filip Jur{\v{c}}{\'\i}{\v{c}}ek, Blaise Thomson, Kai
  Yu, and Steve Young. 2011.
\newblock On-line policy optimisation of spoken dialogue systems via live
  interaction with human subjects.
\newblock In \emph{2011 IEEE Workshop on Automatic Speech Recognition \&
  Understanding}, pages 312--317. IEEE.

\bibitem[{Georgila et~al.(2006)Georgila, Henderson, and
  Lemon}]{georgila2006user}
Kallirroi Georgila, James Henderson, and Oliver Lemon. 2006.
\newblock User simulation for spoken dialogue systems: Learning and evaluation.
\newblock In \emph{Ninth International Conference on Spoken Language
  Processing}.

\bibitem[{Graciarena et~al.(2006)Graciarena, Shriberg, Stolcke, Enos,
  Hirschberg, and Kajarekar}]{graciarena2006combining}
Martin Graciarena, Elizabeth Shriberg, Andreas Stolcke, Frank Enos, Julia
  Hirschberg, and Sachin Kajarekar. 2006.
\newblock Combining prosodic lexical and cepstral systems for deceptive speech
  detection.
\newblock In \emph{2006 IEEE International Conference on Acoustics Speech and
  Signal Processing Proceedings}, volume~1, pages I--I. IEEE.

\bibitem[{He et~al.(2017)He, Balakrishnan, Eric, and
  Liang}]{he-etal-2017-learning}
He~He, Anusha Balakrishnan, Mihail Eric, and Percy Liang. 2017.
\newblock Learning symmetric collaborative dialogue agents with dynamic
  knowledge graph embeddings.
\newblock In \emph{Proceedings of the 55th Annual Meeting of the Association
  for Computational Linguistics (Volume 1: Long Papers)}, pages 1766--1776,
  Vancouver, Canada. Association for Computational Linguistics.

\bibitem[{Ji et~al.(2016)Ji, Liu, He, and Zhao}]{ji2016knowledge}
Guoliang Ji, Kang Liu, Shizhu He, and Jun Zhao. 2016.
\newblock Knowledge graph completion with adaptive sparse transfer matrix.
\newblock In \emph{Thirtieth AAAI Conference on Artificial Intelligence}.

\bibitem[{Krishnamurthy et~al.(2018)Krishnamurthy, Majumder, Poria, and
  Cambria}]{krishnamurthy2018deep}
Gangeshwar Krishnamurthy, Navonil Majumder, Soujanya Poria, and Erik Cambria.
  2018.
\newblock A deep learning approach for multimodal deception detection.
\newblock \emph{arXiv preprint arXiv:1803.00344}.

\bibitem[{Kulkarni et~al.(2016)Kulkarni, Narasimhan, Saeedi, and
  Tenenbaum}]{kulkarni2016hierarchical}
Tejas~D Kulkarni, Karthik Narasimhan, Ardavan Saeedi, and Josh Tenenbaum. 2016.
\newblock Hierarchical deep reinforcement learning: Integrating temporal
  abstraction and intrinsic motivation.
\newblock In \emph{Advances in neural information processing systems}, pages
  3675--3683.

\bibitem[{Levitan et~al.(2018)Levitan, Maredia, and
  Hirschberg}]{levitan2018acoustic}
Sarah~Ita Levitan, Angel Maredia, and Julia Hirschberg. 2018.
\newblock Acoustic-prosodic indicators of deception and trust in interview
  dialogues.
\newblock \emph{Proc. Interspeech 2018}, pages 416--420.

\bibitem[{Li et~al.(2017)Li, Chen, Li, Gao, and Celikyilmaz}]{li2017end}
Xiujun Li, Yun-Nung Chen, Lihong Li, Jianfeng Gao, and Asli Celikyilmaz. 2017.
\newblock End-to-end task-completion neural dialogue systems.
\newblock \emph{arXiv preprint arXiv:1703.01008}.

\bibitem[{Li et~al.(2016)Li, Lipton, Dhingra, Li, Gao, and Chen}]{li2016user}
Xiujun Li, Zachary~C Lipton, Bhuwan Dhingra, Lihong Li, Jianfeng Gao, and
  Yun-Nung Chen. 2016.
\newblock A user simulator for task-completion dialogues.
\newblock \emph{arXiv preprint arXiv:1612.05688}.

\bibitem[{Li et~al.(2015)Li, Tarlow, Brockschmidt, and Zemel}]{li2015gated}
Yujia Li, Daniel Tarlow, Marc Brockschmidt, and Richard Zemel. 2015.
\newblock Gated graph sequence neural networks.
\newblock \emph{arXiv preprint arXiv:1511.05493}.

\bibitem[{Lu et~al.(2005)Lu, Tsechpenakis, Metaxas, Jensen, and
  Kruse}]{lu2005blob}
Shan Lu, Gabriel Tsechpenakis, Dimitris~N Metaxas, Matthew~L Jensen, and John
  Kruse. 2005.
\newblock Blob analysis of the head and hands: A method for deception
  detection.
\newblock In \emph{Proceedings of the 38th Annual Hawaii International
  Conference on System Sciences}, pages 20c--20c. IEEE.

\bibitem[{Meservy et~al.(2005)Meservy, Jensen, Kruse, Burgoon, Nunamaker,
  Twitchell, Tsechpenakis, and Metaxas}]{meservy2005deception}
Thomas~O Meservy, Matthew~L Jensen, John Kruse, Judee~K Burgoon, Jay~F
  Nunamaker, Douglas~P Twitchell, Gabriel Tsechpenakis, and Dimitris~N Metaxas.
  2005.
\newblock Deception detection through automatic, unobtrusive analysis of
  nonverbal behavior.
\newblock \emph{IEEE Intelligent Systems}, 20(5):36--43.

\bibitem[{Niepert et~al.(2016)Niepert, Ahmed, and
  Kutzkov}]{niepert2016learning}
Mathias Niepert, Mohamed Ahmed, and Konstantin Kutzkov. 2016.
\newblock Learning convolutional neural networks for graphs.
\newblock In \emph{International conference on machine learning}, pages
  2014--2023.

\bibitem[{Peng et~al.(2017)Peng, Li, Li, Gao, Celikyilmaz, Lee, and
  Wong}]{peng2017composite}
Baolin Peng, Xiujun Li, Lihong Li, Jianfeng Gao, Asli Celikyilmaz, Sungjin Lee,
  and Kam-Fai Wong. 2017.
\newblock Composite task-completion dialogue policy learning via hierarchical
  deep reinforcement learning.
\newblock In \emph{Proceedings of the 2017 Conference on Empirical Methods in
  Natural Language Processing}, pages 2231--2240.

\bibitem[{P{\'e}rez-Rosas et~al.(2015)P{\'e}rez-Rosas, Abouelenien, Mihalcea,
  Xiao, Linton, and Burzo}]{perez2015verbal}
Ver{\'o}nica P{\'e}rez-Rosas, Mohamed Abouelenien, Rada Mihalcea, Yao Xiao,
  CJ~Linton, and Mihai Burzo. 2015.
\newblock Verbal and nonverbal clues for real-life deception detection.
\newblock In \emph{Proceedings of the 2015 Conference on Empirical Methods in
  Natural Language Processing}, pages 2336--2346.

\bibitem[{Pietquin and Dutoit(2006)}]{pietquin2006probabilistic}
Olivier Pietquin and Thierry Dutoit. 2006.
\newblock A probabilistic framework for dialog simulation and optimal strategy
  learning.
\newblock \emph{IEEE Transactions on Audio, Speech, and Language Processing},
  14(2):589--599.

\bibitem[{Scarselli et~al.(2009)Scarselli, Gori, Tsoi, Hagenbuchner, and
  Monfardini}]{scarselli2009graph}
Franco Scarselli, Marco Gori, Ah~Chung Tsoi, Markus Hagenbuchner, and Gabriele
  Monfardini. 2009.
\newblock The graph neural network model.
\newblock \emph{IEEE Transactions on Neural Networks}, 20(1):61--80.

\bibitem[{Trouillon et~al.(2017)Trouillon, Dance, Gaussier, Welbl, Riedel, and
  Bouchard}]{trouillon2017knowledge}
Th{\'e}o Trouillon, Christopher~R Dance, {\'E}ric Gaussier, Johannes Welbl,
  Sebastian Riedel, and Guillaume Bouchard. 2017.
\newblock Knowledge graph completion via complex tensor factorization.
\newblock \emph{The Journal of Machine Learning Research}, 18(1):4735--4772.

\bibitem[{Vezhnevets et~al.(2017)Vezhnevets, Osindero, Schaul, Heess,
  Jaderberg, Silver, and Kavukcuoglu}]{vezhnevets2017feudal}
Alexander~Sasha Vezhnevets, Simon Osindero, Tom Schaul, Nicolas Heess, Max
  Jaderberg, David Silver, and Koray Kavukcuoglu. 2017.
\newblock Feudal networks for hierarchical reinforcement learning.
\newblock In \emph{Proceedings of the 34th International Conference on Machine
  Learning-Volume 70}, pages 3540--3549. JMLR. org.

\bibitem[{Wang et~al.(2019)Wang, Zhang, Li, Hwang, Zong, and
  Li}]{wang-etal-2019-incremental}
Weikang Wang, Jiajun Zhang, Qian Li, Mei-Yuh Hwang, Chengqing Zong, and Zhifei
  Li. 2019.
\newblock Incremental learning from scratch for task-oriented dialogue systems.
\newblock In \emph{Proceedings of the 57th Annual Meeting of the Association
  for Computational Linguistics}, pages 3710--3720, Florence, Italy.
  Association for Computational Linguistics.

\bibitem[{Wang et~al.(2018)Wang, Zhang, Zhang, Hwang, Zong, and
  Li}]{wang-etal-2018-teacher}
Weikang Wang, Jiajun Zhang, Han Zhang, Mei-Yuh Hwang, Chengqing Zong, and
  Zhifei Li. 2018.
\newblock A teacher-student framework for maintainable dialog manager.
\newblock In \emph{Proceedings of the 2018 Conference on Empirical Methods in
  Natural Language Processing}, pages 3803--3812, Brussels, Belgium.
  Association for Computational Linguistics.

\bibitem[{Wen et~al.(2017)Wen, Vandyke, Mrk{\v{s}}i{\'c}, Gasic, Barahona, Su,
  Ultes, and Young}]{wen2017network}
Tsung-Hsien Wen, David Vandyke, Nikola Mrk{\v{s}}i{\'c}, Milica Gasic, Lina
  M~Rojas Barahona, Pei-Hao Su, Stefan Ultes, and Steve Young. 2017.
\newblock A network-based end-to-end trainable task-oriented dialogue system.
\newblock In \emph{Proceedings of the 15th Conference of the European Chapter
  of the Association for Computational Linguistics: Volume 1, Long Papers},
  volume~1, pages 438--449.

\bibitem[{Williams et~al.(2017)Williams, Asadi, and Zweig}]{williams2017hybrid}
Jason~D Williams, Kavosh Asadi, and Geoffrey Zweig. 2017.
\newblock Hybrid code networks: practical and efficient end-to-end dialog
  control with supervised and reinforcement learning.
\newblock In \emph{Proceedings of the 55th Annual Meeting of the Association
  for Computational Linguistics (Volume 1: Long Papers)}, volume~1, pages
  665--677.

\bibitem[{Williams(1992)}]{williams1992simple}
Ronald~J Williams. 1992.
\newblock Simple statistical gradient-following algorithms for connectionist
  reinforcement learning.
\newblock \emph{Machine learning}, 8(3-4):229--256.

\bibitem[{Young et~al.(2013)Young, Ga{\v{s}}i{\'c}, Thomson, and
  Williams}]{young2013pomdp}
Steve Young, Milica Ga{\v{s}}i{\'c}, Blaise Thomson, and Jason~D Williams.
  2013.
\newblock Pomdp-based statistical spoken dialog systems: A review.
\newblock \emph{Proceedings of the IEEE}, 101(5):1160--1179.

\end{thebibliography}
\bibliographystyle{acl_natbib}

\end{document}